\documentclass[sigconf]{acmart}

\usepackage{booktabs} % For formal tables
\usepackage{url}
\setcopyright{rightsretained}
\acmDOI{}

\acmISBN{}

\author{Farhan Khawar}
\affiliation{\institution{The Hong Kong University of Science and Technology}}
\email{fkhawar@cse.ust.hk}

\author{Nevin L. Zhang}
\affiliation{\institution{The Hong Kong University of Science and Technology}}
\email{lzhang@cse.ust.hk}

\begin{document}
\title{Matrix Factorization Equals Efficient Co-occurrence Representation}

\begin{abstract}
Matrix factorization is a simple and effective solution to the recommendation problem. It has been extensively employed in the industry and has attracted much attention from the academia. However, it is unclear what the low-dimensional matrices represent. We show that matrix factorization can actually be seen as simultaneously calculating the eigenvectors of the user-user and item-item sample co-occurrence matrices. We then use insights from random matrix theory (RMT) to show that picking the top eigenvectors corresponds to removing sampling noise from user/item co-occurrence matrices. Therefore, the low-dimension matrices represent a reduced noise user and item co-occurrence space. We also analyze the structure of the top eigenvector and show that it corresponds to global effects and removing it results in less popular items being recommended. This increases the diversity of the items recommended without affecting the accuracy.

 \end{abstract}


\maketitle

\section{Introduction}
Matrix Factorization (MF) is a preferred collaborative filtering (CF) method due to its speed, effectiveness, and ease of deployment. It works by taking a user-item data matrix $\mathbf{X}$ and factorizing it into a product of two low-rank ($k$) matrices i.e. $\mathbf{X} = \mathbf{U}{\mathbf{V}}{^T}$.

Truncated singular value decomposition (SVD) is the best reduced-rank approximation of $\mathbf{X}$ under the $L^2$ norm.
Various MF methods that currently exits for CF \cite{Hu:2008:CFI:1510528.1511352} can be seen as adaptations and extensions of SVD by introducing regularization and unequal weights in the loss function. However, unlike these methods, SVD provides an analytically analyzable solution which will help us gain insights into the working of these methods.

The SVD of $\mathbf{X} = \mathbf{U\Sigma}\mathbf{V}^T$, where the matrix $\mathbf{U}$ contains the left singular vectors, the matrix $\mathbf{V}$ contains the right singular vectors and the diagonal matrix $\mathbf{\Sigma}$ contains the singular values of $\mathbf{X}$. We know that the columns of $\mathbf{U}$ and $\mathbf{V}$ are the \emph{eigenvectors} of $\mathbf{X}\mathbf{X}^T$ and $\mathbf{X}^T\mathbf{X}$ respectively. In addition, $\mathbf{X}^T\mathbf{X}$ and $\mathbf{X}\mathbf{X}^T$ share the same eigenvalues which are equal to the square root of the singular values in $\mathbf{\Sigma}$. If we notice that $\mathbf{X}\mathbf{X}^T$ is the user-user co-occurrence matrix and $\mathbf{X}^T\mathbf{X}$ is the item-item co-occurrence matrix, then by performing the singular value decomposition of $\mathbf{X}$, we are actually calculating the spectrum of the user-user co-occurrence matrix and the item-item co-occurrence matrix simultaneously. Therefore, MF at its heart operates on user-user and item-item correlations and fuses these two pieces of information.

\section{Sampling Noise}
Since, $\mathbf{X}$ is a sample of the true user-item consumption matrix, $\mathbf{X}^T\mathbf{X}$ and $\mathbf{X}\mathbf{X}^T$ are also sample co-occurrence matrices that contain sampling noise. We illustrate this with the aid of the Mar$\check{c}$enko Pastur law (MP-law)\cite{marvcenko1967distribution}. It applies to the case where an $n  \times m$  random matrix $\mathbf{X}$ is large, $m,n \rightarrow \infty$, but the number of samples is not too large i.e. the ratio $m/n \rightarrow q \in (0,1]$\footnote{A similar result is derived for $q>1$}.

Under these conditions the eigenvalues distribution of the covariance matrix of $\mathbf{X}$ i.e., $ \mathbf{C}=\frac{1}{n}\mathbf{X}^T\mathbf{X}$ is known exactly and is given by the Mar$\check{c}$enko Pastur law:
\begin{equation}
\label{RMT}
\rho_{\mathbf{X}}(\lambda) = \frac{1}{2 \pi q \lambda} \sqrt{(\lambda_{max}-\lambda)(\lambda-\lambda_{min})},
\end{equation}
where the eigenvalue $\lambda \in [\lambda_{max}, \lambda_{min}] $ and $\lambda_{max} = (1+\sqrt{q})^2$ and $\lambda_{min} = (1-\sqrt{q})^2$.

\begin{figure}
		\centering
		\includegraphics[width=3.6in, height = 1.5in]{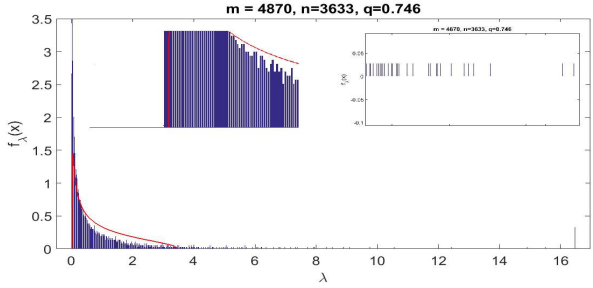}
		\vspace{-7mm}
		\caption{The solid line shows the plot of the theoretical MP-law density from (\ref{RMT}). The histogram shows the eigenvalue density of the Movielens1M data. The eigenvalues in the noise bulk are zoomed in the top left and the eigenvalues outside the bulk are shown on the top right.}
		\label{rand}
\end{figure}

The true eigenvalue density of a random matrix is a spike at 1, however, the MP-law states that due to the finite sampling size the eigenvalue density spreads according to (\ref{RMT}). This spread is known as ''noise bulk''. A plot of the density of (\ref{RMT}) is shown in Fig. \ref{rand} along with the eigenvalue distribution of the Movielens1M item correlation matrix. Eigenvalues inside the noise bulk are indistinguishable from noise and the eigenvalues outside signify a prominent signal.

Since CF datasets also fall in the finite sample regime, therefore, the user-user and item-item co-occurrence matrices also suffer from the sampling noise which manifests as eigenvalue spreading. MP-law suggests a natural way to mitigate this noise i.e., taking the top $k$ eigenvalues and associated eigenvectors. We can then interpret the low-rank matrices obtained as the reduced noise eigenvectors that span the user/item co-occurrence space. Thus, taking the top eigenvalues and associated eigenvectors of the co-occurrence matrices has the interpretation of mitigating the noise ingrained in the co-occurrence matrices.

\section{The highest eigenvalue}
\begin{figure}
		\centering
		\includegraphics[scale=0.4]{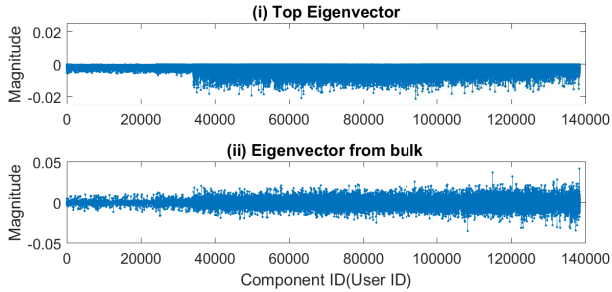}
		\caption{Components of $\mathbf{v}_{H_u}$ and and eigenvector from the noise bulk.}
		\vspace{-6mm}
		\label{topEV}
\end{figure}

Given the premise that reduced-rank MF under $L^2$ norm can be viewed as an eigenvalue problem, we can extract some insights from the eigenvectors of the co-occurrence. In this work, we focus eigenvector, $\mathbf{v}_{H}$, which is associated with the highest magnitude eigenvalue. This eigenvector represents global effects of the system which cause all users or all items to co-occur. Examples of such effects can be users who consume a lot of items and, conversely, popular items that tend to be consumed by most users. Fig. \ref{topEV} shows the plot of the components of $\mathbf{v}_{H_u}$ for $\mathbf{X}\mathbf{X}^T$ . It can be seen that a global factor, like popular items, is causing all users to ''move'' together and be correlated. This is in contrast to the noise bulk eigenvector where no apparent correlation was observed.
%\footnote{The top eigenvector, $\mathbf{v}_{H_i}$, of $\mathbf{X}^T\mathbf{X}$ has a similar structure.}

\subsection{Experiments: Effect of $\mathbf{v}_H$}
The resultant effect of multiplying $\mathbf{v}_{H_i}$ and $\mathbf{v}_{H_u}$ in the MF recommender would be to promote popular items and the choices of popular users. To investigate this, we performed experiments on the Movilens20M dataset. The rating magnitudes were ignored to binarize the dataset. In addition, results and chosen parameters are based on 5-fold cross validation (CV). Three scenarios were tested: (a) retaining only the top $k$ eigenvectors ; (b) retaining top $k$ eigenvectors except $\mathbf{v}_{H}$\footnote{Removing both $\mathbf{v}_{H_u}$ and $\mathbf{v}_{H_i}$ by setting $\Sigma_{1,1}=0$}; and (c) retaining only the top eigenvector $\mathbf{v}_{H}$ of $\mathbf{X}\mathbf{X}^T$ and $\mathbf{X}^T\mathbf{X}$. The results are shown in Table \ref{tab.res}.
%\footnote{ \url{https://grouplens.org/datasets/movielens/20m/}.}
It can be seen that the accuracy (NDCG@50 and recall@50) for case (a) and (b) is almost the same, thus it is not affected by the removal of $\mathbf{v}_H$. However, the diversity measured by the number of unique items recommended to all users (D@50) changes appreciably.% This is a desirable outcome.

\begin{figure}
\centering
		\includegraphics[scale=0.35]{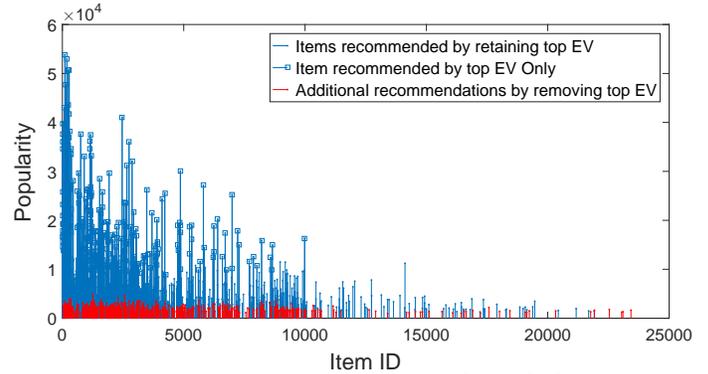}
		\vspace*{-8mm} 
		\caption{Removing $\mathbf{v}_H$ increases diversity by including non-popular items.}
		\label{remTop}
\end{figure}

To investigate which type of new items are recommended by removing $\mathbf{v}_H$, in Fig. \ref{remTop} we plotted the popularity of the items recommended under the three scenarios. The items in blue correspond to the recommendations in scenario (a). Among these items, the ones with marked by a square are the 290 items recommended by scenario (c). Finally, the items recommended in scenario (b) are in red and blue, where the items in red are the additional items recommended if we remove the global effect represented by $\mathbf{v}_H$. We see that these additional items are non-popular items, thus signifying that the effect of removing $\mathbf{v}_H$ is increased diversity by recommending non-popular items. We also see that $\mathbf{v}_H$ encourages recommending popular items only, as in scenario (c) only items above 10,000 views are recommended.

\begin{table}[]
\centering
\vspace{-5mm}
\caption{Accuracy, global diversity and CV time results.}
\label{tab.res}
\small
\begin{tabular}{lllll}
\hline
\textbf{Method}                     & \textbf{NDCG@50} & \textbf{Recall@50} & \textbf{D@50}   & \textbf{Time}(min.) \\
\hline
\textbf{(a)}SVD\tiny{$(k=20)$ }             & 0.60597 & 0.40434   & 1574 & 34.8       \\
\textbf{(b)}SVD\tiny{$(k=19)$}              & 0.60168 & 0.40088   & 2139 & 35.4       \\
\textbf{(c)}SVD\tiny{$(k=1)$ }             & 0.42106 & 0.19704   & 290  & 20.8       \\
\hline
SVD\tiny{$(k=100)$}             & 0.59912 & 0.37539   & 2368   & 88       \\
WRMF\tiny{$(k=20,\lambda=10^-3)$} & 0.60678 & 0.40904   & 1861.6 & 214     \\
\hline

\end{tabular}
\end{table}

\subsection{Experiments: Other results}
Other conclusions from the lower half of Table \ref{tab.res} are that if the number of eigenvalues retained is too large( e.g. SVD$(k=100)$) then noise can result in decreased accuracy. Also, SVD$(k=19)$ performs very similar to WRMF \cite{Hu:2008:CFI:1510528.1511352}, but with increased diversity and much lower running time. The lower running time is due to the linear complexity of the efficient truncated SVD solvers like Lanczos bidiagonalization \citep{baglama2005augmented}.

\section{Future Work}
Future directions to explore include: analyzing the structure of other eigenvectors outside the noise bulk; the effect of removing $\mathbf{v}_H$ in terms of popular users; and the effect of standardizing $\mathbf{X}$, so that eigenvectors correspond to the covariance matrix of $\mathbf{X}$.

Research on this article was supported by Hong Kong Research Grants Council under grant 16202118.
\vspace{-1.5mm}
\bibliographystyle{ACM-Reference-Format}
\bibliography{MF} 

\end{document}